\documentclass{article}

\PassOptionsToPackage{round, sort&compress}{natbib}

\usepackage[preprint]{nips_2018}

\usepackage[utf8]{inputenc} 
\usepackage[T1]{fontenc}    
\usepackage{hyperref}       
\usepackage{url}            
\usepackage{booktabs}       
\usepackage{amsfonts}       
\usepackage{nicefrac}       
\usepackage{microtype}      
\usepackage{graphicx}

\usepackage[draft=true]{minted}
\usepackage{authblk}
\usepackage{caption}

\title{DELTA \\
\large A DEep learning based Language Technology plAtform}

\author{Kun Han}
\author{Junwen Chen}
\author[]{Hui Zhang}
\author[]{Haiyang Xu}
\author[]{Yiping Peng}
\author[]{Yun Wang}
\author[]{Ning Ding}
\author[]{Hui Deng}
\author[]{Yonghu Gao}
\author[]{Tingwei Guo}
\author[]{Yi Zhang}
\author[]{Yahao He}
\author[]{Baochang Ma}
\author[]{Yulong Zhou}
\author[]{Kangli Zhang}
\author[]{Chao Liu}
\author[]{Ying Lyu}
\author[]{Chenxi Wang}
\author[]{Cheng Gong}
\author[]{Yunbo Wang}
\author[]{Wei Zou}
\author[]{Hui Song}
\author[]{Xiangang Li}
\affil{AI Labs, DiDi Chuxing\thanks{Correspondence to: Kun Han, Junwen Chen, and Hui Zhang: \{kunhan,chenjunwen, ethanzhanghui\}@didiglobal.com}}

\begin{document}

\maketitle

\begin{abstract}
    In this paper we present DELTA, a deep learning based language technology platform. DELTA is an end-to-end platform designed to solve industry level natural language and speech processing problems. It integrates most popular neural network models for training as well as comprehensive deployment tools for production. DELTA aims to provide easy and fast experiences for using, deploying, and developing natural language processing and speech models for both academia and industry use cases. We demonstrate the reliable performance with DELTA on several natural language processing and speech tasks, including text classification, named entity recognition, natural language inference, speech recognition, speaker verification, etc. DELTA has been used for developing several state-of-the-art algorithms for publications and delivering real production to serve millions of users.
\end{abstract}

\section{Introduction}
In recent years, deep learning has been achieving tremendous success on numerous machine learning applications \citep{lecun2015deep,schmidhuber2015deep}. It firstly dramatically improved the state-of-the-art in speech recognition \citep{hinton2012deep} and image recognition \citep{krizhevsky2012imagenet}, and then produced extremely promising results on natural language processing (NLP) \citep{collobert2011natural,sutskever2014sequence,devlin2019bert}.

In natural language and speech processing areas, various algorithms and models are emerging recently due to the rapid progress in both academia and industry. Recently, many open-sourced libraries are released for community development since the interest in applying deep learning approaches to NLP and speech processing is very high. However, some open-sourced libraries are implemented for specific tasks, and researchers and engineers have to adapt the code for their own problems. More importantly, most existing implementations are designed for research purposes and do not consider model serving. Although TensorFlow provides interfaces for model serving, it is not trivial to deploy the models for complex production usage.

We introduce DELTA,\footnote{https://github.com/didi/delta} a deep learning based natural language and speech processing platform, as an end-to-end open-sourced library, which integrates most popular NLP and speech processing models and deployment tools, forming with uniform code structures and input/output (I/O) interfaces. DELTA is implemented using TensorFlow \citep{tensorflow2015-whitepaper}, featuring easy-to-use, easy-to-deploy, and easy-to-develop:
\begin{itemize}
    \item Easy to use
    \begin{itemize}
        \item It supports most common NLP and speech tasks that works out of the box. The users may directly complete model training through one line of command.
        \item It is of highly-customized configuration. The users can easily customize model architectures and parameters.
        \item It supports multimodal training using textual, acoustic, and numeric features, which is very useful for the multimodal scenarios especially in real applications.
        \item It optimizes the front-end data preprocessing and parallelizes model training so it is easy to train models on a huge amount of data.
    \end{itemize}

    \item Easy to deploy
    \begin{itemize}
        \item What you see in training is what you get in serving. All data processing and feature extraction components are implemented as TensorFlow operators (OPs) and integrated into a graph for deployment.
        \item All models have the uniform I/O interfaces and transparent to outside. One can deploy a new model with no changes on the serving code.
    \end{itemize}

    \item Easy to develop
    \begin{itemize}
        \item The models and components are modularized. The developers can easily build new models on top of them.
        \item All modules are fully-tested and provide reliable and efficient performance.
    \end{itemize}
\end{itemize}

\section{Platform architecture}
This section describes the overall architecture and code structures for DELTA. In high-level, DELTA includes two parts:

\begin{itemize}
    \item Modeling: This is the main package for offline modeling, including data processing, model building, training, and evaluation, etc. The code is organized in \mintinline{python}{delta} directory. We discuss the details for this part in this section.

    \item Deployment: This is the package for online serving using existing model. DELTA provides flexible online serving approaches for various production environments such as GPU machines, mobile devices, embedded devices. The code is organized in \mintinline{python}{deltann} directory. We will discuss this part in the next section.
\end{itemize}

\begin{figure}[htbp]
  \centering
  \includegraphics[width=1.0\linewidth]{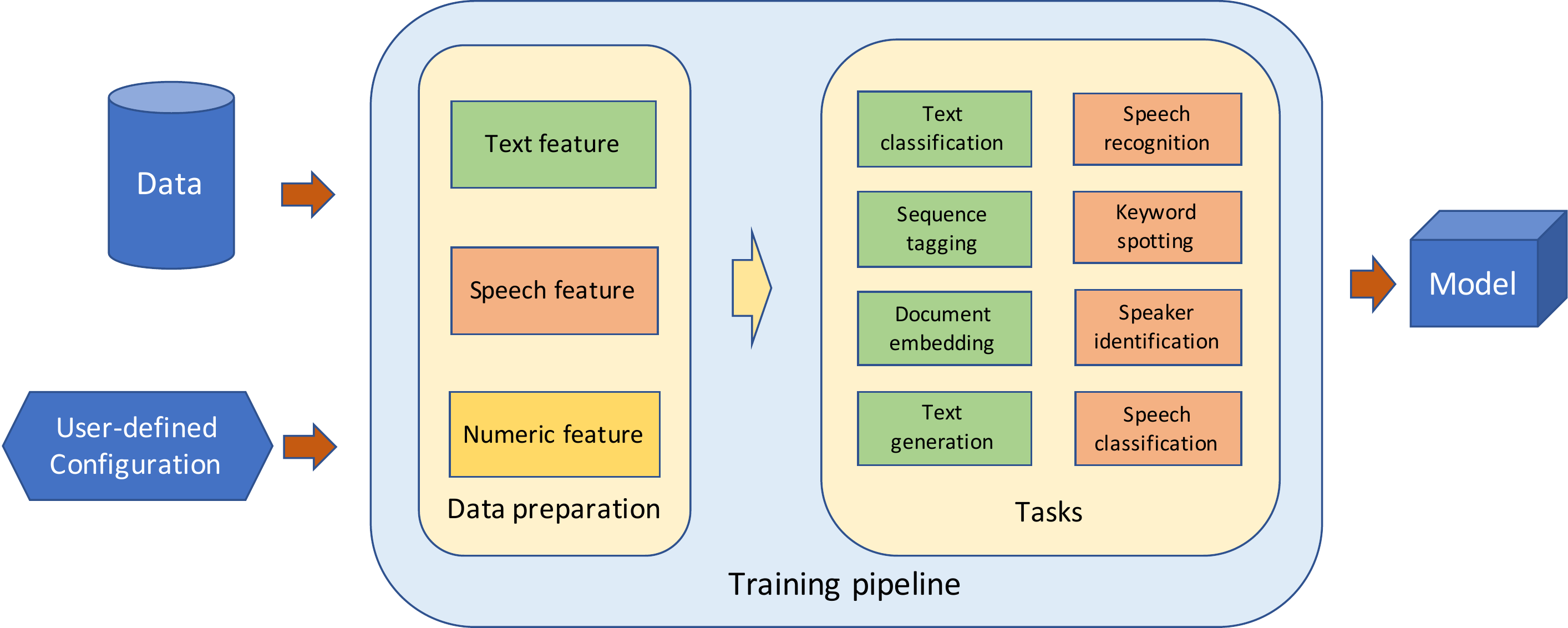}
  \caption{High-level overview of DELTA.}
  \label{fig:highlevel_delta}
\end{figure}

Fig. \ref{fig:highlevel_delta} shows a high-level overview for the modeling part in DELTA. An user is responsible for providing data and a user-defined configuration. Based on the configuration, DELTA can automatically generate a training pipeline including the corresponding data processing and modeling modules. The pipeline will start model training processing and generate a model once complete.

\subsection{Data processing} \label{sec:data}
Acoustic and textual information are the two most important ways for human communication, both of which are presented as sequences but in different modalities. Therefore most sequential models are applicable to both text and speech. In DELTA, we use \mintinline{python}{Dataset} API to load and process data, and the code is modularized in the directory \mintinline{python}{delta/data}\footnote{https://www.tensorflow.org/api\_docs/python/tf/data/Dataset}.

\subsubsection{Text data}

DELTA supports text inputs in the raw text format. Preprocessing transforms the text inputs from the original format to a model-readable format. For English, we provide standard text processing functions using TensorFlow operations, including splitting, punctuation conversion, casing, etc. For Chinese, word segmentation is usually used to organize Chinese characters into words. We implement several Chinese word segmentation algorithms as TensorFlow OPs, such as \textit{jieba}\footnote{https://github.com/fxsjy/jieba}. A \textit{tokenizer} is then applied to transform a text string into a list of indices. In DELTA, \mintinline{python}{SentenceToIdsOp} is applied to index the tokens based on a vocabulary look-up table. All these processing operations are implemented as customized OPs, and some are introduced from the third-party libraries such as \textit{Lingvo}\footnote{https://github.com/tensorflow/lingvo} \citep{shen2019lingvo}. The OPs are organized in \mintinline{python}{delta/layers/ops/py_x_ops.py} and can be compiled as a dynamic library.

\subsubsection{Speech data}

Speech tasks, unlike NLP tasks, usually come with complex feature extraction on raw speech signals. TensorFlow has provided several functions for speech processing, such as MFCC and spectrogram. In DELTA, we also implement speech processing operations for broader applications, including filter-bank feature, perceptual linear prediction feature (PLP), zero crossing rate, frame power, pitch, first order and seconder order derivatives between frames (i.e., deltas and accelerations). To perform high-level transformation, we implemented \mintinline{python}{AnalysisFilterBank} and \mintinline{python}{SynthesisFilterBank} to generate spectrogram and phase features from waveforms and reconstruct waveforms from both, respectively. We reuse some OPs from speech processing libraries. All these OPs are also included in \mintinline{python}{delta/layers/ops/py_x_ops.py} and can be used to either build the model graph or process data.

\subsubsection{Numeric data}
In addition to textual and acoustic features, the numeric feature (a.k.a., dense feature) is commonly used in machine learning. In fact, the numeric feature is a generic feature representation which can be directly fed into the model and most features are converted into numeric features for model training. In practise, a user may need to train a model using numeric features from different domains, e.g., click-through-rate, user profile, etc. We implement model training with numeric features in DELTA, which can be used in general machine learning tasks.


\subsubsection{Multimodal data}

We also want to highlight that DELTA supports multimodal training from different data sources. Multimodal training is very common in real applications. For example, in search engine, both textual features and user historic features (numeric features) are helpful for produce accurate search results. One may need to build a multimodal model to learn from textual features and numeric features simultaneously. Another example is that, in human speech emotion recognition, both acoustic features and textual features can be combined together to build a multimodal emotion recognizer \citep{xu2019alignment}.

To utilize multimodal data, there are many approaches to combine the features \citep{ngiam2011multimodal}. In DELTA, we use direct concatenation to combine the features due to the simplicity. Our implementation includes multimodal training for textual+numeric features, acoustic+numeric features and textual+acoustic features. A user only need to define the training mode in the configuration file to start a multimodal training.

\subsection{Models}

To build models in DELTA, we wrap the low-level neural network models as basic components, which can be directly used in high-level models. The basic components include multi-layer perceptrons (MLPs) \citep{rosenblatt1958perceptron,rumelhart1988learning}, convolutional neural networks (CNNs) \citep{lecun1990handwritten}, long short-term memory (LSTM) \citep{hochreiter1997long}, attention mechanism \citep{bahdanau2014neural}, transformers \citep{vaswani2017attention}, etc. These components are widely used in NLP models and speech models for different tasks and they are implemented in \mintinline{shell}{delta/layers} directory.

Note that, we have implemented the most popular high-level models in DELTA, meaning that a user can directly use the provided configuration files for model training. For most high-level models, our implementation refers to publications, and some modification may be applied. If you want to develop a new high-level model, you build a new model in \mintinline{python}{delta/models} with a python register.

\subsubsection{NLP models}
In DELTA, we organize the NLP models based on the input/output data format. For different NLP tasks, if the input and output data formats are the same, they usually can be addressed by similar high-level architecture. For example, in high-level, a sequence-to-sequence architecture can be used to address both machine translation and text summarization, with some modification for each task. We currently implement the following model architectures:

\begin{itemize}
    \item \textit{Sequence classification}\\
    This model deals with tasks like sentence classification and document classification. Here, the input is a sequence of text and the target is a label. For document classification, you can define a sentence separator to divide the documents into a sequence of sentences for hierarchical modeling. We provide the sequence classification models using CNN \citep{kim2014convolutional,collobert2011natural}, LSTM \citep{tai2015improved}, hierarchical attention networks \citep{yang2016hierarchical}, transformer \citep{vaswani2017attention}, etc.

    \item \textit{Sequence labeling}\\
    For this model, both the input and the output are sequences and the model assigns a label to each step in the input sequence, so the input sequence and the output sequence should have the same length. This model is usually used in named entity recognition (NER), part-of-speech tagging, etc. We have provided the LSTM based sequence labeling \citep{chiu2016named} and an LSTM with CRF based method \citep{huang2015bidirectional,ma2016end,lample2016neural}.

    \item \textit{Pairwise modeling}\\
    The input for this model is a pair of two documents. The model is trained to learn the relationship between them. In this model, each document is modeled separately to generate its document embedding, then the two embedding vectors are either concatenated together and fed into a classifier or compared to compute the cosine similarity. This type of models has been widely used on document similarity, query-based question-answering (QA), natural language inference (NLI), etc. The key for this approach is to supervisedly train the networks for document embedding \citep{huang2013learning,bowman2015large,conneau-EtAl:2017:EMNLP2017}.

    \item \textit{Sequence-to-sequence (seq2seq) modeling} \\
    The seq2seq model takes input as a sequence of text and generate a sequence of text. But the lengths of two sequences are not necessarily the same. This structure usually has an encoder to learn the information from the input sequence and then uses a decoder to autoregressively generate the output. It has been widely used in many text generation tasks, for example, machine translation, text summarization, dialogue generation, etc. We implement the standard seq2seq models using LSTM with attention \citep{bahdanau2014neural} and transformers \citep{vaswani2017attention}. Note that, the seq2seq structure is also used for speech recognition. In DELTA, this part is shared between NLP and ASR tasks.

    \item \textit{Multi-task modeling} \\
    Since the models in DELTA are modularized, it is easy to support multi-task learning using the existing modules. As an example, we implement a multi-task model for sequence classification and labeling, where the sequence level loss and the step level loss are computed simultaneously. This model is used to jointly train an intent recognizer and named entity recognizer together. (Junwen add more details and citations)

    \item \textit{Pretraining integration} \\
    Recent NLP studies have been showing significant improvements on using pretrained language models on unlabeled data, e.g., ELMO \citep{Peters:2018}, BERT \citep{devlin2019bert}. We implement an interface to integrate a pretrained model into a DELTA model, where the pretrained model is used to dynamically generate embedding which is concatenated with the word embedding for the different task. To be specific, a user can pretrain an ELMO or BERT model first and then build a DELTA model with the prertained model. Both model will be combined into a TensorFlow graph for training and inference. The ELMO or BERT models trained from the official open-sourced libraries can be directly used in DELTA.

\end{itemize}

\subsubsection{Speech models}
For speech processing, the model usually couple with the feature extraction because different task may have different features extracted using signal processing. In order to better organize the model, we implement the speech models as a task-specific fashion.

\begin{itemize}
    \item \textit{Automatic speech recognition (ASR)}\\
    ASR is one of the most important task in speech processing, where the input is a raw speech waveform in time series and the output is the corresponding text. Recent trend in ASR is to build an end-to-end ASR system using seq2seq framework. We provide an attention based seq2seq ASR model \citep{chan2016listen,bahdanau2016end}. We also implement another popular type of ASR model using connectionist temporal classification (CTC) \citep{graves2006connectionist}.


    \item \textit{Speaker verification}\\
    Speaker verification is the process of verifying whether an utterance belongs to a specific speaker, based on that speaker's known utterances (i.e., enrollment utterances). It has applications on voice match, identification check, etc. We provide an X-vector text-independent model \citep{snyder2018xvector} and an end-to-end model \citep{wan2017ge2e}.


    \item \textit{Speech emotion recognition}\\
    From the machine learning perspective, speech emotion recognition is similar to speaker identification, because both are speech sentence classification. However, the features used in both tasks can be very different, therefore we provide both models separately. Recently several deep learning based approaches have been successfully used in speech emotion recognition \citep{han2014speech,satt2017efficient} and we implement some models the in DELTA.




\end{itemize}

\subsubsection{Multimodal models}
Multimodal modeling is paramount to practical applications because real data usually comes from different sources and available in multiple modalities. Since it is still a research topic to how to model multimodal data, we provide a straightforward approach to concatenate the multimodal data in the input feature stage for modeling \citep{ngiam2011multimodal}. This is a framework to use multimodal data for general purpose. More sophisticated data fusion approaches are usually task dependent and we provide an example using aligned textual and acoustic features for emotion recognition \citep{xu2019alignment}.

\begin{itemize}
    \item \textit{Textual+acoustic}\\
    A common applications for multimodal learning with text and speech is speech emotion recognition, where the model takes input as speech signals and the corresponding transcripts and output the emotion states. In our implementation, we use two sequential models (e.g., CNNs or LSTMs) to learn the sequence embedding for speech and text separately, and then concatenates the learned embedding vectors for classification.

    \item \textit{Textual+numeric}\\
    Multimodal learning with textual and numeric features was widely used in real application. In DELTA, we implement the direct concatenation data fusion in data processing stage, therefore this type of multimodal training can be directly used for existing models in DELTA.
\end{itemize}

\subsection{Training pipeline}

\begin{figure}[htbp]
  \centering
  \includegraphics[width=1.0\linewidth]{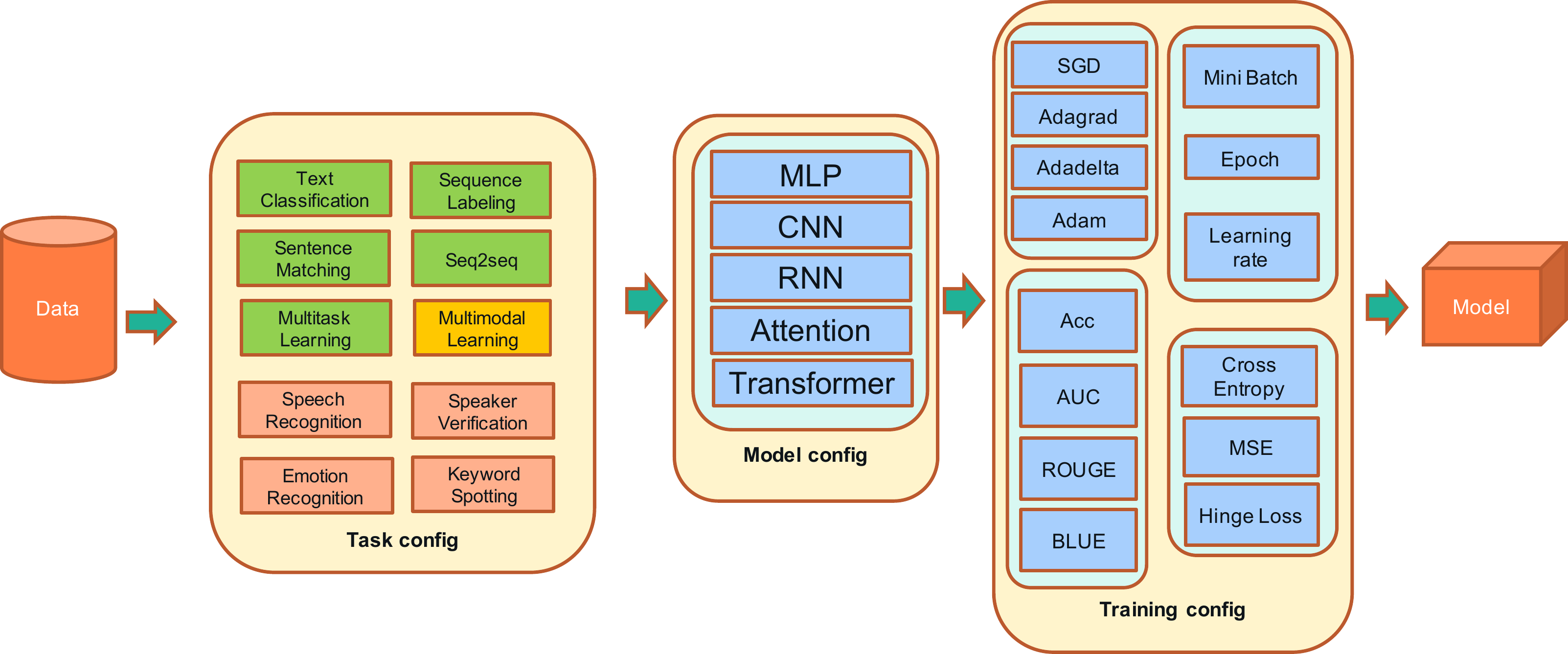}
  \caption{Configuration of model training in DELTA.}
  \label{fig:config}
\end{figure}

In this subsection, we introduce the whole training pipeline in DELTA. With the existing models in DELTA, a user only need to prepare the dataset and provide a configuration file in the format of yaml. As shown in Fig. \ref{fig:config}, the configuration contains the parameters used in model training. The task configuration indicates the model or task types including several NLP and speech tasks as well as the multimodal tasks. The model configuration defines the model parameters such as types of neural networks, number of hidden layers, number of neural units. The training configuration mainly specify the parameters for optimization engines and evaluation metrics. Once the configuration files are defined, you can simply start a training pipeline with one line of command.

To deep dive into the training pipeline, we now discuss the detailed implementation methods in DETLA. From the aspect of code structure, a model training process uses a uniform training pipeline, which consists of three components: \textit{task}, \textit{model}, and \textit{solver}.

\subsubsection{Task}

\textit{Task} is an abstract class for data processing as mentioned in Section \ref{sec:data}, including feature processing and data formatting. It applies online or offline data processing, and the processed data can be fed into model for training, evaluation, or inference. It includes several abstract methods for generating, batching, feature preprocessing and shuffling the data examples. After data processing, it wraps the data examples by \mintinline{python}{tf.data.dataset}, which supports several functions such as  batching, padding, shuffling, multi-threading processing. This is similar to the design in Tensor2Tensor \citep{tensor2tensor} and compatible with both TensorFlow 1.x and 2.x. The abstract methods in \textit{Task} are brief described as follows:

    \begin{itemize}
        \item \mintinline{python}{generate_data} is a python generator to yield one example in every iteration.

        \item \mintinline{python}{feature_spec} is a specification of the \textit{types} and \textit{shapes} of the data example generated by \mintinline{python}{generate_data}.

        \item \mintinline{python}{preprocess_batch} is the main class for feature processing of batched examples. You can apply different feature processing in this class, such as \textit{text tokenization}, \textit{Chinese word segmentation}, \textit{audio feature extraction}, \textit{speech feature normalization}. Note that, in DELTA, each feature preprocessing operations have been implemented as a TensorFlow OP, therefore it is straightforward to use it as a model component and build an end-to-end model graph. To develop new feature processing OP, one can also program customized TensorFlow OPs in DELTA.

        \item \mintinline{python}{dataset} is the main class of data pipeline, and a wrapper for \mintinline{python}{tf.data.dataset}. All data processing of examples are organized in this class.

        \item \mintinline{python}{input_fn} generates a function without parameters, which is also recommended by TensorFlow Estimator\footnote{https://www.tensorflow.org/guide/estimators}.
    \end{itemize}

    For speech processing tasks, e.g., automatic speech recognition and speaker identification, the inputs are usually raw waveform files and you may need to apply extra offline feature processing, such as speech feature extraction and global feature normalization. These methods mainly follow the feature processing part in the speech recognition library Kaldi \citep{Povey_ASRU2011}.

    \begin{itemize}
        \item \mintinline{python}{generate_feat} is responsible for generating speech feature from raw speech data, such as spectrogram feature or Mel-frequency cepstral coefficients (MFCCs).
        \item \mintinline{python}{generate_cmvn} compute global cepstral mean and variance normalization (CMVN) on generated speech feature.
    \end{itemize}

\subsubsection{Model building}
For model building, to compatible with both TensorFlow 1.x and 2.x, we introduce two types of models in DELTA: \textit{Keras model} and \textit{raw model}. Keras \citep{chollet2015keras} is a popular high-level API for building deep learning models, which has been adopted into TensorFlow 2.x, where the high-level API is changed from \mintinline{python}{tf.layers} to \mintinline{python}{tf.keras.layers}. To build a new model in DELTA, we recommend using Keras API because it is user-friendly and easy to extend. To compatible with legacy system, DELTA also supports the raw model type as in TensorFlow 1.x, which can be deployed into production for TensorFlow 1.x system.

    \begin{itemize}
        \item \textit{Keras model} \\
        Keras models supports object-oriented design, including variants, layers, models and checkpoint management. This is the default model type in DELTA and we use the subclass style paradigm of Keras to construct sublayers or submodels. In DETLA, a model is initialized in \mintinline{python}{__init__()} method and called in \mintinline{python}{call()} method. 

        \item \textit{Raw model} \\
        For the sake of compatibility, we also use raw models. The interface of raw models is similar to that of Keras models, so we use the unified paradigm for model building. We use \mintinline{python}{tf.variable_scope} to wrapper the scope of the model implementation.
    \end{itemize}

\subsubsection{Solver}
    A \textit{Solver} is a module for composing the pipeline containing model construction, training and evaluation process. When building a model, you can choose a solver based on the model or task types. For example, you may want to choose \mintinline{python}{RawClassSolver} if the model you build is a subclass of \mintinline{python}{TextClassModel}. The most recommended solver base clase are \mintinline{python}{EstimatorSolver} and \mintinline{python}{RawSolver}. The main abstract methods of \textit{Solver} are described as follows:

    \begin{itemize}
        \item \mintinline{shell}{train()} builds a model for the training mode. This process include training data fetching, iteration on datasets, optimization, and parameters updating. The process is usually implemented with \mintinline{python}{tf.train.MonitoredTrainingSession}. This function saves the model to a TensorFlow checkpoint for every specified training steps.

        \item \mintinline{shell}{eval()} builds the model for the evaluation mode, including evaluation data fetching with labels, inference, and measurement. The metrics are defined in \mintinline{python}{delta.utils.metrics}.

        \item \mintinline{shell}{infer()} is similar to \mintinline{shell}{eval()}, but it takes unlabeled data as inputs for inference. In addition, for some tasks, e.g., keyword spotting \citep{chen2014kws}, we need to add post processing after the model inference, so \mintinline{shell}{infer()} runs model inference followed by a \mintinline{shell}{postproc_fn()} function.

        \item \mintinline{shell}{train_and_eval()} is the combination of \mintinline{shell}{train()} and \mintinline{shell}{eval()}. It applies the evaluation for every specified training steps.

        \item \mintinline{shell}{export_model()} transforms the model checkpoint saved during training process to a TensorFlow \textit{SavedModel} format\footnote{https://www.tensorflow.org/guide/saved\_model}, which is supported by most deployment environments.
    \end{itemize}

\section{Deployment}
To our knowledge, most open-sourced deep learning platforms are designed for model training and do not focus on model serving. However, serving a model for industrial applications is not trivial. In order to facilitate productizing a trained model, we provide deployment features for model serving in DELTA.

\begin{figure}[htbp]
  \centering
  \includegraphics[width=1.0\linewidth]{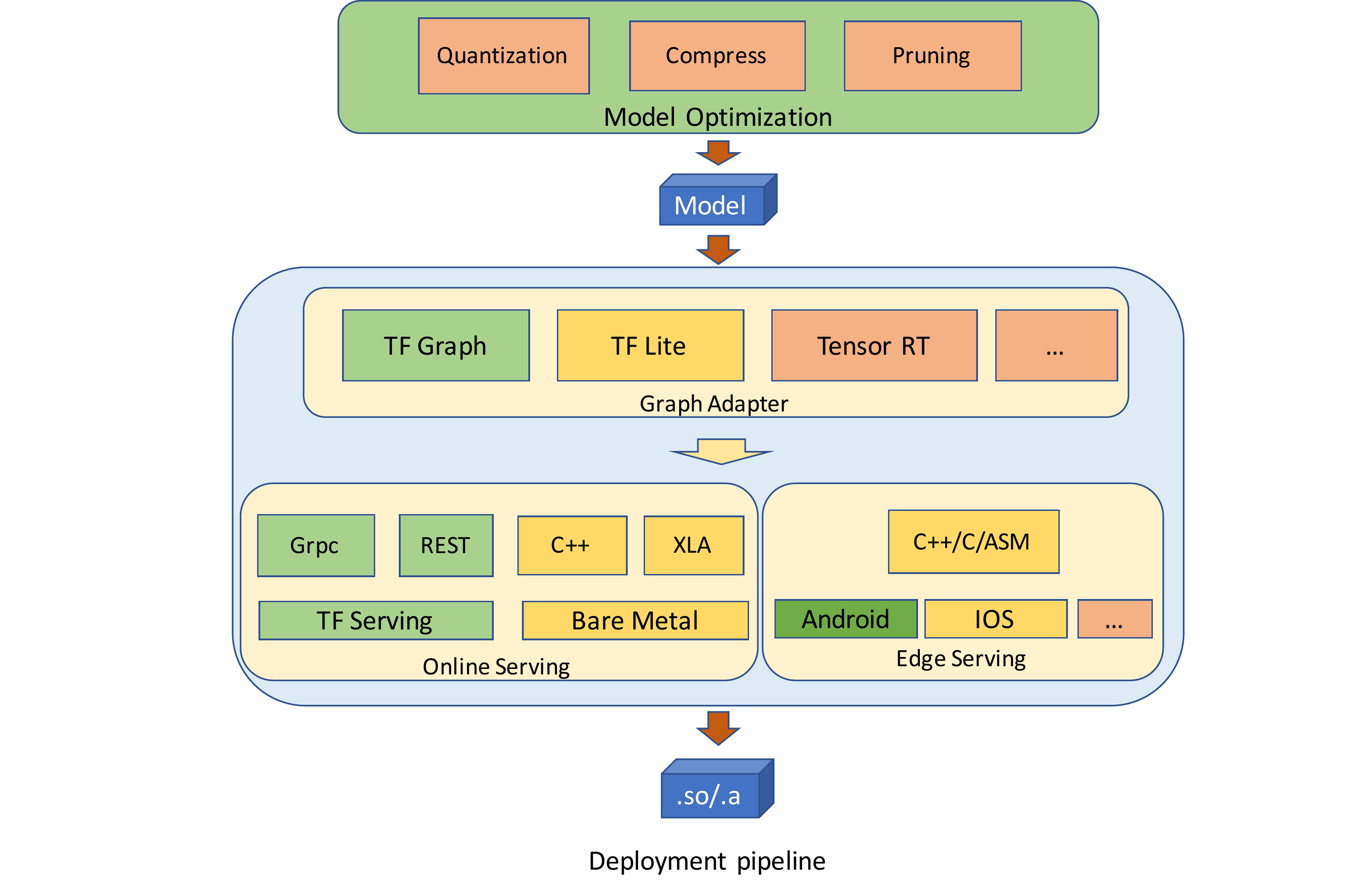}
  \caption{High-level overview of productization.}
  \label{fig:highlevel_deltann}
\end{figure}

The deployment part is referred as DELTA-NN and implementations are mainly organized under \mintinline{python}{deltnn} \mintinline{python}{dpl}. Fig. \ref{fig:highlevel_deltann} shows an overview of DELTA-NN. A trained model is optionally compressed or optimized in terms of computational efficiency. The deployment pipeline is able to convert the model into several formats as needed. As shown in the figure, TF Serving supports two interfaces, GRPC and RESTful API. They are widely used in online serving in production. For a bare metal environment, you can wrap DELTA-NN as a library with interfaces for model inference.

\subsection{Model Optimization}
Optimizing the computational efficiency is important to real applications. Deploying model into mobile or edge devices has restricted requirement on the models, such as limited memory and low power consumption. Some embedded devices do not have floating-point units and can only support integer accelerator. In DELTA-NN, we integrate a model distillation component which is used to learn a small model from an existing large model \citep{hinton2015distilling}. provide an interface for model optimization, such as model quantization, model pruning. We also

\subsection{Model serving}

For productization, DELTA-NN supports multi-graphs online and/or offline inference, with CPU, GPU and edge devices. There are several serving modes: TFModel, TFLite and TFServing. We can also combine these serving modes as a customized mode.

\begin{itemize}
    \item \textit{TFModel} \\
    This mode is provided by the standard TensorFlow library. TensorFlow provides several model formating, including CheckPoint, GraphDef, and SavedModel, etc. We recommend using SavedModel as the model format for standard model serving.




    \item \textit{TFLite} \\
    TensorFlow Lite\footnote{https://www.tensorflow.org/lite} is a lightweight solution for mobile and embedded devices for TensorFlow models. It enables low-latency inference of on-device machine learning models with a small binary size and fast performance supporting hardware acceleration. DELTA-NN supports using TensorFlow Lite as an engine for mobile and embedded devices.

    \item \textit{TFServing} \\
    TensorFlow Serving\footnote{https://github.com/tensorflow/serving} is a flexible, high-performance serving system for machine learning models, designed for production environments. It deals with the inference aspect of machine learning, taking models after training and managing their lifetimes, providing clients with versioned access via a high-performance, reference-counted look-up table. TensorFlow Serving provides out-of-the-box integration with TensorFlow models, and can be easily extended to serve other types of models. DELTA-NN implements an HTTP/HTTPS Client to request remote TensorFlow Serving model results. Combining with TFLite, you can easily deploy offline serving solutions.

\end{itemize}

\subsection{Deployment pipeline}
The deployment pipeline is organized as scripts and put into the folder \mintinline{shell}{dpl}.

A user is responsible for providing TensorFlow \textit{SavedModel} and a user-defined \textit{model configuration}. Based on the configuration, DELTA-NN can automatically generate a deployment pipeline, including the following steps:

\begin{enumerate}
    \item Model optimization  \\
    This is a place-holder for model distillation, quantization, pruning, compression. You can choose to use a third-party library for model optimization.


    \item Model graph transformation \\
    This step converts a \textit{SavedModel} to a proper model format for serving on cloud, mobile and/or embedded devices, e.g., frozen graph, saved model, TFLite model, etc. Scripts are under \mintinline{shell}{dpl/gadpter}.

    \item Compiling library \\
    This step compiles the model inference engine library. It will compile models on different platforms and devices with the corresponding engine libraries, e.g., TensorFlow C++ library for CPU and GPU, TFLite library for Android and iOS, etc.


    After compiling the engine library, the step will compile DELTA-NN library. DELTA-NN abstracts low-level engine interface as a uniform interface, which supports multi-graphs inference,  offline and online mixed inference, and multiple devices inference. These features are very useful in production.

    The engine library and DELTA-NN library are then combined into a composed binary for application.

    \item Testing  \\
    Testing is critical to an industrial level application. We release a Docker image\footnote{https://www.docker.com/} to perform unit test, integration test and system test. This process is automated through continuous integration (CI).

    A user can perform A/B test as part of CI. You can set up some metrics as trigger signals for continuous deployment when deploying a new model. We recommend using Docker to maintain the deployment process.

\end{enumerate}

\section{Benchmarks}
In order to evaluate the performance of the released models in DELTA, we provide experimental results for each task on public datasets as benchmarks. Note that, the purpose for the experiments is to provide solid and reliable benchmarks, therefore most experiments use those commonly-used models rather than the state-of-the-art algorithms. For each task, we compare our model with a similar model chosen from a highly-cited publication.

In DELTA, we organize all experiment settings together with the data processing scripts in the directory \mintinline{shell}{egs}, similar to the usage in Kaldi \citep{Povey_ASRU2011} or ESPnet \citep{watanabe2018espnet}. As a DELTA user, you can easily reproduce the experimental results by using the configure files under corresponding dataset directories. You can also extend the algorithms based on these models for better performance.

\subsection{NLP benchmark}
We present the comparison results of our implementation on NLP models. For each task, we choose a model from a popular reference as the baseline. We use DELTA to build the similar models as in the publications and use the same datasets and metrics for comparison. The baseline results are the number provided in the reference. Table~\ref{nlp-performance-table} shows the comparison results on several different NLP tasks and our results are on a par with the baselines.

\begin{minipage}{\textwidth}
  \captionof{table}{Performances of NLP models}
  \label{nlp-performance-table}
  \centering
  \small
  \begin{tabular}{lllllll}
    \toprule
    Task & Model & Dataset & Metric & DELTA & Baseline & Reference \\
    \toprule
    \multicolumn{2}{l}{\textit{Sequence classification}} & & & & \\
    Sentences  & CNN & TREC~\footnote{https://trec.nist.gov/data.html} & Acc & 92.2 & 91.2 & \cite{kim2014convolutional} \\
    Documents  & HAN & Yahoo Answer~\footnote{https://webscope.sandbox.yahoo.com/\#datasets} & Acc & 75.1 & 75.8 & \cite{yang2016hierarchical}  \\
    \midrule
    \multicolumn{2}{l}{\textit{Sequence labeling}} & & & & \\
    NER & BLSTM-CRF & CoNLL2003~\footnote{https://www.clips.uantwerpen.be/conll2003/ner/} & F1 & 84.6 & 84.7 & \cite{huang2015bidirectional}\\
    \midrule
    \textit{Multitask}  & & & & \\
    Intent & BLSTM-CRF & ATIS~\footnote{https://catalog.ldc.upenn.edu/docs/LDC93S4B/corpus.html} & Acc & 97.4 & 98.2 & \cite{liu2016attention}\\
    Slot filling &  &  & F1 & 95.2 & 95.9 & \\
    \midrule
    \multicolumn{2}{l}{\textit{Pairwise modeling}}  & & & & \\
    NLI & LSTM & SNLI~\footnote{https://nlp.stanford.edu/projects/snli/} & Acc & 80.7 & 80.6 & \cite{bowman2016fast} \\
    \midrule
    \textit{Seq2seq}  & & & & \\
    Summarization & BLSTM & CNN/Daily Mail~\footnote{https://github.com/deepmind/rc-data} & RougeL & 27.3 & 28.1~\footnote{Our model is slightly different from that of the reference, so a small performance gap is expected.} & \cite{see2017get}\\
    \midrule
    \multicolumn{2}{l}{\textit{Pretrained model}}  & & & & \\
    NER & ELMO & CoNLL2003 & F1 & 92.2 & 92.2 & \cite{Peters:2018}\\
    NER & BERT & CoNLL2003 & F1 & 94.6 & 94.9 & \cite{devlin2019bert} \\
    \bottomrule
  \end{tabular}
\end{minipage}

\subsection{Speech benchmark}
We are working on benchmarks for speech tasks, and we will update as soon as available.

\section{Conclusion}
We present DELTA, a deep learning based NLP and speech processing platform. DELTA provides modularized components and simple pipelines for use and develop. It also contains many deployment supports for real production. We conducted experiments on public datasets to demonstrate the reliable and solid implementation.

For future work, we will implement more advanced models and present the state-of-the-art results for NLP and speech tasks using DELTA. We also interested in the implementation of parallel model training, automatic parameter tuning, production integration, etc. More importantly, DELTA is an open-sourced library and the contribution from the researchers and engineers in the community is highly appreciated.


\bibliographystyle{plainnat}
\bibliography{ref_delta}

\end{document}